\documentclass[journal]{IEEEtran}
\usepackage{cite}

\ifCLASSINFOpdf
   \usepackage[pdftex]{graphicx}
   \DeclareGraphicsExtensions{.pdf,.jpeg,.png,.eps}
\else
   \usepackage[dvips]{graphicx}
   \DeclareGraphicsExtensions{.eps}
\fi
\usepackage{amsmath}
\usepackage{amsthm}
\usepackage{amssymb}
\usepackage{algorithmic}
\usepackage{algorithm}

\usepackage{array}

\ifCLASSOPTIONcompsoc
  \usepackage[caption=false,font=footnotesize,labelfont=sf,textfont=sf]{subfig}
\else
  \usepackage[caption=false,font=footnotesize]{subfig}
\fi
 \usepackage{dblfloatfix}

\ifCLASSOPTIONcaptionsoff
  \usepackage[nomarkers]{endfloat}
 \let\MYoriglatexcaption\caption
 \renewcommand{\caption}[2][\relax]{\MYoriglatexcaption[#2]{#2}}
\fi
\usepackage{url}

\begin{document}
%
\title{Multimodal Deep Network Embedding with Integrated Structure and Attribute Information}
%
%
%
%

\author{Conghui~Zheng,
        Li~Pan,
        and~Peng~Wu
\thanks{This work is supported by National Key Research and Development Plan in China (2017YFB0803300), National Natural Science Foundation of China (U1636105). \textit{(Corresponding author: Li Pan.)}}
\thanks{C. Zheng, L. Pan and P. Wu are with the School of Electronic Information and Electrical Engineering, Shanghai Jiao Tong University, Shanghai, CN, 200240.
(e-mail: \{chzheng, panli, catking\}@sjtu.edu.cn.)}%
\thanks{Manuscript received Jun 15, 2018; revised Dec 11, 2018.}}

%
%

\markboth{Journal of \LaTeX\ Class Files,~Vol.~14, No.~8, June~2018}%
{Zheng \MakeLowercase{\textit{et al.}}: Multimodal Deep Network Embedding with Integrated Structure and Attribute Information}
%



\maketitle

\begin{abstract}
Network embedding is the process of learning low-dimensional representations for nodes in a network, while preserving node features. Existing studies only leverage network structure information and focus on preserving structural features. However, nodes in real-world networks often have a rich set of attributes providing extra semantic information. It has been demonstrated that both structural and attribute features are important for network analysis tasks. To preserve both features, we investigate the problem of integrating structure and attribute information to perform network embedding and propose a Multimodal Deep Network Embedding (MDNE) method. MDNE captures the non-linear network structures and the complex interactions among structures and attributes, using a deep model consisting of multiple layers of non-linear functions. Since structures and attributes are two different types of information, a multimodal learning method is adopted to pre-process them and help the model to better capture the correlations between node structure and attribute information. We employ both structural proximity and attribute proximity in the loss function to preserve the respective features and the representations are obtained by minimizing the loss function. Results of extensive experiments on four real-world datasets show that the proposed method performs significantly better than baselines on a variety of tasks, which demonstrate the effectiveness and generality of our method.
\end{abstract}


%
\IEEEpeerreviewmaketitle

\section{Introduction}
\label{sec:1}
%
%
%
%
\IEEEPARstart{N}{etwork} mining is the basis for many network analysis tasks, such as classification and link prediction. The dimensionality of traditional node representations is proportional to the network scale, which requires large amount of storage and computation resources for network analysis tasks. Thus, it is necessary to learn low-dimensional representations of nodes to capture and preserve the network features. Network embedding, also known as network representation learning, is a way of learning low-dimensional representations and preserving useful features to commonly support subsequent network analysis tasks.

Existing network embedding methods \cite{perozzi2014deepwalk,grover2016node2vec,cao2015grarep} that emphasize preserving network structural features have achieved promising performance in several network analysis tasks. However, nodes in real-world networks have rich attribute information beyond the structural details, such as text information in citation networks and user profiles in social networks. Attribute features are essential to network analysis applications \cite{hu2013exploiting,tang2013exploiting}and it is insufficient to learn network representations only based on preserving structural features. Node attributes carry semantic information that largely alleviates the link sparsity problem and supplement the incompleteness structure information. The strong correlations between structures and attributes enable them to be integrated to learn network representations according to the principles of homophily \cite{mcpherson2001birds} and social influence theory \cite{marsden1988homogeneity}. Therefore, we integrate the topological structures and node attributes to perform network embedding to preserve both structural and attribute features of the network. Differing from some task-oriented network embedding methods that learn network representations for the specific task, we aim to learn the network representations generally applying to various advanced network analysis tasks. We face three challenges: (1) Both the underlying network structures \cite{luo2011cauchy} and the complex interactions between attributes and structures \cite{cui2017survey} are highly non-linear. Thus, designing a model to capture these non-linear relationships is difficult. (2) The structures and attributes are information from different sources, which make it difficult to find direct correlations between originally observed information due to sparsity and noise. Modeling the correlations between network structures and attribute information is a tough problem. (3) The nodes with coherent links and similar attributes in the original network have strong proximity. They are supposed to be close to each other in the embedding space as well. Thus, mapping the proximity of nodes from both structure and attribute perspectives to the embedding space is critically important.

To address the above challenges, a Multimodal Deep Network Embedding method named MDNE is proposed in this paper. Most of existing shallow models have limited ability to represent complex non-linear relationships \cite{bengio2009learning}. A deep model comprising of multiple layers of non-linear functions, using each layer to capture the non-linear relationships of units in the lower layer, is able to extract the non-linear relationships of data progressively during training \cite{hinton2006reducing}. Moreover, deep learning has been demonstrated to have powerful non-linear representation and generalization ability \cite{bengio2009learning}. In order to capture the highly non-linear network structures and the complex interactions between structures and attributes, a deep model comprising multiple layers of non-linear functions is proposed to learn compact representations of nodes.The original structure and attribute information, which are represented by an adjacency matrix and attribute matrix, respectively, are usually sparse and noisy, making it difficult for the deep model to extract the correlations between them directly. In this paper, a multimodal learning method \cite{ngiam2011multimodal} is adopted to pre-process the structure and attribute information to obtain their high-order features. High-order features are condensed and less noisy, so concatenating the two high-order features facilitates the deep model to extract the high-order correlations between the network structures and node attributes. To ensure the obtained representations preserve both structural and attribute features of the original network, we use the structural proximity and attribute proximity to define the loss function for the new model. We preserve the structural features by taking the advantage of the first-order proximity and second-order proximity, which capture the local and global network structures \cite{wang2016structural}. The attribute proximity, which indicates the similarity of node attributes, is also utilized in the learning process to preserve the attribute features of the network. Thus, the learned representations preserve both the structural and attribute features of nodes in the embedding space.

To evaluate the effectiveness and generality of the proposed method in a variety of scenes, we conduct experiments to analyze network representations obtained by different network embedding methods from four real-world network datasets in three analysis tasks including link prediction, attribute prediction, and classification. The results show that the network representations obtained by MDNE offer better performance on different tasks compared to other methods. This demonstrates that the proposed method effectively preserves the topological structure and attribute features of nodes in the embedding space, which improves the performance on diverse network analysis tasks.

The rest of the paper is organized as follows. Section 2 discusses the related works. The proposed method MDNE is described in details in Section 3. Experimental results of different network analysis tasks on various real-world datasets are presented in Section 4. Finally, Section 5 concludes the paper.

\section{Related Works}
\label{sec:2}
The early works of network embedding are related to graph embedding \cite{roweis2000nonlinear,belkin2002laplacian}, which aims to embed an affinity graph into a low-dimensional vector space. The affinity graph is obtained by calculating the proximity between feature vectors of nodes. Recent network embedding aims to embed naturally formed networks into a low-dimensional space, such as social networks, citation networks, etc. Most of the existing works \cite{jacob2014learning,hofmann2001unsupervised,blei2003latent} focused on reducing the dimensions of structure information while preserving the structural features of nodes. GraRep \cite{cao2015grarep} defined different loss functions of models to preserve high-order proximity among nodes and optimized each model by matrix factorization techniques. The final representations of nodes combined representations learned from different models. M-NMF \cite{wang2017community} proposed a novel modularized nonnegative matrix factorization model to incorporate the community structure into network embedding. The above shallow models have applied in various network analysis tasks, but have limited ability to represent the highly non-linear structure of networks. Thus, techniques with deep models were introduced to deal with the problem. LINE \cite{tang2015line} designed the objective function based on the first-order proximity and second-order proximity and adopted negative sampling approach to minimize the objective function to get low-dimensional representations which preserve the local and global structure of the network. DeepWalk \cite{perozzi2014deepwalk} utilized random walks in the network to sample the neighbors of nodes. By regarding the path generated as sentences, it adopted Skip-Gram, a general word representation learning model, to learn the node representations. Node2vec \cite{grover2016node2vec} modified the way of generating node sequences and proposed a flexible notion of node's neighborhood. A biased random walk procedure was designed, which explored diverse neighborhood. SDNE \cite{wang2016structural} designed a clear objective function to preserve the first-order proximity and second-order proximity of nodes and mapped the network into a highly non-linear latent space through an autoencoder-based model.

Besides structure information, most of the recently obtained network datasets often carry a large amount of attribute information. However, it is difficult for pure structure-based methods to compress attribute information and obtain the representations combining the structure and attribute information. Therefore, efforts have been done to jointly exploit structure and attribute information in network embedding, and the representations integrating structure and attribute information have been demonstrated to improve the performance in network analysis tasks \cite{hu2013exploiting,tang2013exploiting,li2016robust}. TADW \cite{yang2015network} proved DeepWalk to be equivalent to matrix factorization and incorporated text features into network representation learning under the framework of matrix factorization. It can only handle the text attributes. AANE \cite{huang2017accelerated} modeled and incorporated node attribute proximity into network embedding in a distributed way. The above matrix factorization methods did not preserve the attribute features directly, but performed the learning based on the attribute affinity matrix calculated by a specific affinity metric, which limited the attribute feature preservation ability of the obtained representations. UPP-SNE \cite{zhang2017user} learned joint embedding representations by performing a non-linear mapping on user profiles guided by network structure. It mainly dealt with user profile information. TriDNR \cite{pan2016tri} separately learned embedding from a coupled neural network architecture and linearly combined them in an iterative way. It lacked sufficient knowledge interactions between the two separate models. ASNE \cite{Liao2017Attributed} proposed a multilayer perceptron framework to integrate the structural and attribute features of nodes. It preserved the structural proximity and attribute proximity by maximizing the likelihood function defined based on random walks. Its model lacked a non-linear pre-processing of the structure and attribute information, which could facilitate to extract the high-order correlations between attribute and structural features in the later learning. In this paper, a multimodal learning method is adopted to pre-process the original data.

Multimodal learning methods which have aroused considerable research interests aim to project data from multiple modalities into a latent space. The classical methods CCA, PLS, BLM and their variants \cite{hardoon2004canonical,rosipal2005overview,tenenbaum2000separating} were widely applied in previous time. Recent decades have seen great power of deep learning method to generate integrated representations for multimodal data. \cite{ngiam2011multimodal} proposed an autoencoder-based method to learn features over multiple modalities (video and audio) and achieved in speech recognition. \cite{srivastava2012learning} proposed a Deep Belief Network (DBN) architecture for learning a joint representation of multimodal data, which made it possible to create representations when some data modalities are missing. The multimodal Deep Boltzmann Machine (DBM) model proposed in  \cite{srivastava2014multimodal} fused modalities (image and tag) together and extracted unified representations which were useful for classification and information retrieval tasks. \cite{kang2015learning} learned consistent representations for two modalities and facilitated the cross-matching problem. \cite{xu2017learning} proposed a cross-modal hashing method to learn unified binary representations for multimodal data. Following these successful works, we introduced multimodal learning method into network embedding. The structure and attribute information of the network are regarded as different modalities. An autoencoder-based multimodal model \cite{ngiam2011multimodal} is adopted to pre-process the bimodal data and forms high-order features, which facilitate the fused representations to be learnt.

There are also methods learning network representations for specific applications. PinSage \cite{ying2018graph} combined the recent Graph Convolutional Network (GCN) algorithm \cite{defferrard2016convolutional,kipf2016semi} with efficient random walks to generate representations applying in web-scale recommender systems. However, our MDNE learns integrated representations generally applying for various network analysis tasks.

\section{Multimodal Deep Network Embedding Method}
\label{sec:3}
\subsection{Problem Definition}
An attributed network is defined as $G=(U,E,A)$, where $U=\{u_1,\dots,u_n\}$ represents a set of $n$ nodes, $E=\{e_{i,j}\}$ represents a set of $l$ edges, and $A=\{\bf{a}_i\}_{i=1}^n$ represents the attribute matrix. Edge information is represented by the adjacency matrix $S=\{\bf{s}_i\}_{i=1}^n$.

The adjacency vector $\bf{s}_i$ and attribute vector $\bf{a}_i$ of node $i$ represent the structure and attribute information, respectively. Thus, the goal of network embedding is to compress the two vectors into a low-dimensional vector, and preserving the structure and attribute features in the low-dimensional space (embedding space).

The first-order proximity and second-order proximity capture the local and global network structural features, respectively \cite{wang2016structural}.
\newtheorem{definition}{Definition}
\begin{definition}[First-Order Structural Proximity]
  The first-order proximity describes the local pairwise proximity between two nodes. For each pair of nodes, the edge weight, $s_{i,j}$ indicates the first-order proximity between $u_i$ and $u_j$.
\end{definition}
\begin{definition}[Second-Order Structural Proximity]
  The second-order proximity between a pair of nodes $(u_i,u_j)$ in a network describes the similarity between their neighborhood structures which are represented by the adjacency vectors.
\end{definition}
 The first-order proximity and second-order proximity jointly compose the structural proximity between nodes. The attribute proximity captures the attribute feature of nodes.
\begin{definition}[Attribute Proximity]
  The attribute proximity between a pair of nodes $(u_i,u_j)$ describes the proximity of their attributes information. It is determined by the similarity between their attribute vectors, i.e., $a_i$ and $a_j$.
\end{definition}
The attribute proximity and structural proximity between nodes are the basis of many network analysis tasks. For example, community detection on social networks clusters nodes based on the structural proximity and attribute proximity \cite{wu2018mining}. In recommendation on citation networks, papers having strong structural and attribute proximity are most likely to be reference papers of the given manuscript \cite{cai2018three}. In user alignment across social networks, users are aligned based on their structure and attribute proximity on each network \cite{zhao2018learning}. These applications benefit from utilizing both structural proximity and attribute proximity, which lead us to vestigate the problem of learning the low-dimensional representations of the network in the condition of preserving the two proximities. The problem is defined as follows.
\begin{definition}[Attributed Network Embedding]
  Given an attributed network denoted as $G=(U,E,A)$ with $n$ nodes and $m$ attributes, attributed network embedding aims to learn a mapping function $f:(\bf{s}_i,\bf{a}_i)\mapsto \bf{y}_i\in \mathbb{R}^d$, where $d \ll \min (n,m)$. The objective of the function is to make the similarity between ${{\bf{y}}_{i}}$ and ${{\bf{y}}_{j}}$ explicitly preserve the attribute proximity and structural proximity of ${u_i}$ and ${u_j}$.
\end{definition}

\subsection{Framework}
\label{sec:3.2}
In order to address the attributed network embedding problem, a Multimodal Deep Network Embedding (MDNE) method is proposed. Figure 1 shows the MDNE framework. The parameters marked with $\hat{}$ are parameters of the reconstruction component. Table 1 lists the  terms and notations. Note that the attributes pre-processing layer and structures pre-processing layer have different weight matrices ${W^{{a^{(1)}}}}$ and ${W^{{s^{(1)}}}}$, respectively. For simplicity, we denote ${W^{{a^{(1)}}}}$ and ${W^{{s^{(1)}}}}$ as ${W^{(1)}}$.

The strong interactions and complex dependencies between nodes in real-world networks result in the high non-linearity of the network structures. The interactions between structure and attribute features are non-linear as well. Deep neural networks have demonstrably strong representation and generalization abilities for such non-linear relationships \cite{he2016deep}. Therefore, the proposed model is established based on a deep autoencoder, one of the most common deep neural network architectures. Autoencoder is an unsupervised learning model that performs well in data dimensionality reduction and feature extraction \cite{hinton2006reducing}. An autoencoder consists of two parts, the encoder and decoder. The encoder consists of one or multiple layers of non-linear functions that map the input data into the representation space and obtain its feature vector; the decoder reconstructs the data in the representation space to obtain its original input form by an inverse process. A shallow autoencoder has three layers (input, encoding and output), where the encoder has only one layer of non-linear functions. The deep autoencoder of our implementation has more hidden layer and is able to learn higher-order features of data. Given the input data vector $\bf{x}_i$, the output feature vectors for each layer are
\begin{equation*}
\begin{array}{l}
{\bf{y}_i}^{(1)} = \sigma \left( {{W^{(1)}}{\bf{x}_i} + {\bf{b}^{(1)}}} \right)\\
{\bf{y}_i}^{\left( k \right)} = \sigma \left( {{W^{(k)}}{\bf{y}_i}^{(k - 1)} + {\bf{b}^{(k)}}} \right),k = 2,\dots,K
\end{array}
\end{equation*}
, where $\sigma$ denotes the non-linear activation function for bringing the non-linearity into the models. The activation functions must be chosen according to the loss function \cite{charte2018practical}, the requirements of the applied representations, and the datasets. In practice, we can choose them based on their test performance. In this work, the sigmoid function $\sigma (x) = \frac{1}{{1 + \exp ( - x)}}$ is adopted as it provided the best performance in the experiments\footnote{Regarding the choice of activation function, we have tried sigmoid, Rectified Linear Unit (ReLU), Scaled Exponential Linear Unit (SELU), and hyperbolic tangent function (tanh). Empirically, the sigmoid function leads to the best performance in general.}. After obtaining the mid-layer representation, i.e., the encoding result ${\bf{y}_i}^{(K)}$, we can obtain the decoding result through an inverse calculation process. The autoencoder optimizes the parameters by minimizing the reconstruction error between the input data and the reconstructed data. A typical loss function is the mean squared error (MSE)
\begin{equation*}
{\cal L} = \sum\limits_{i = 1}^n {\left\| {{{\bf{\hat x}}_i} - {\bf{x}_i}} \right\|_2^2}
\end{equation*}
. To alleviate the noise and redundant information in the input feature vectors, an undercomplete autoencoder is adopted to learn compact low-dimensional representations. The undercomplete autoencoder has a tower structure, with each upper layer having a smaller number of neurons than the layer below it. A smaller number of neurons restricts the dimensionality of the learned features, so that the autoencoder is forced to learn more abstract features of data during training \cite{charte2018practical}. A layer-by-layer pre-training algorithm, such as Restricted Boltzmann Machine (RBM) enables each upper layer of the encoder to capture the high-order correlations between the feature units in the lower layer, which is an efficient way to extract non-linear structures  progressively \cite{hinton2006reducing}. Thus, the tower structure with stacked multiple layers of non-linear functions is able to map the data into a compressive latent space, and capture the highly non-linear structures of the network, as along with the complex interactions between the structures and attributes during training. The basic undercomplete autoencoder is chosen in our framework because of its generality and simplicity. Variants of the autoencoder can replace the basic autoencoder with slight modifications to accommodate specific scenarios, such as denoising autoencoder, contractive autoencoder, etc. \cite{charte2018practical}.
\begin{figure}[!t]
\centering
\includegraphics{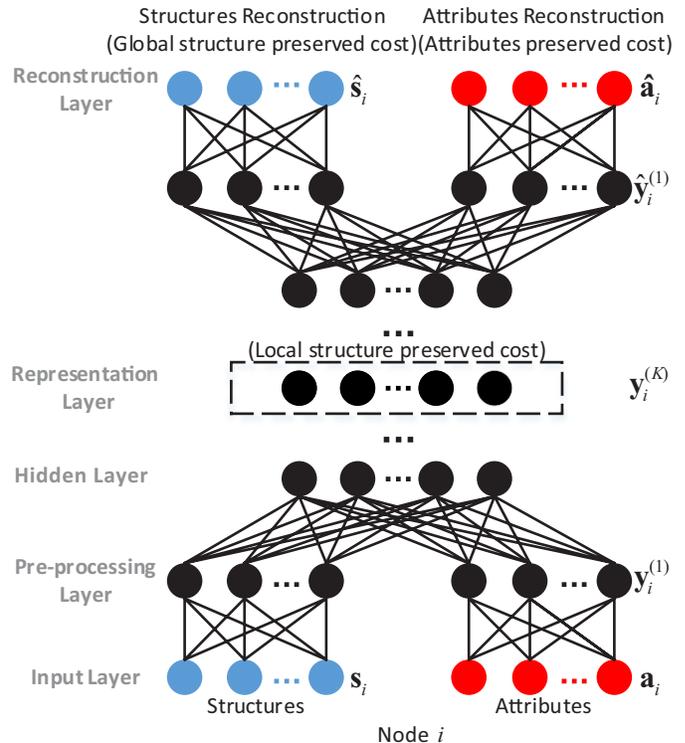}
\caption{The framework of MDNE}
\label{F1}
\end{figure}
\begin{table}[!t]
\renewcommand{\arraystretch}{1.3}
\caption{Terms and notations}
\label{T1}
\centering
\begin{tabular}{|c|c|}
\hline
Symbol & Definition   \\
\hline
$K$ & Number of layers of the encoder/decoder \\
\hline
${W^{(k)}},{\hat W^{(k)}}$ & Weight matrix of the ${k^{th}}$ layer \\
\hline
${{\bf{b}}^{(k)}},{{\bf{\hat b}}^{(k)}}$ & Biases of the ${k^{th}}$ layer \\
\hline
${Y^{(k)}} = \{ {{\bf{y}}_i}^{(k)}\} _{i = 1}^n$ & Representations of the ${k^{th}}$ layer \\
\hline
\end{tabular}
\end{table}

An intuitive way to integrate both structure and attribute information in the representations is to concatenate the two feature vectors separately learned from both modalities. The way of learning individual modalities separately is limited in its ability to extract the correlations between structures and attributes. Alternatively, two kinds of information can be concatenated first at the input and the integrated representations are learned by a unified model. The inputs of the unified model are the adjacency vectors describing network structure and the attribute vectors describing node attributes. Since the adjacency vectors and attribute vectors of nodes are sparse and noisy, inputting the concatenated adjacency vector and attribute vector to the deep autoencoder directly, as shown in Figure 2(a), increases the difficulty in training the model to capture the correlations between structure and attribute information. We have also found that, in practice, learning in this way results in hidden units have strong connections of either structure or attribute variables, but few units connect across the two modalities \cite{ngiam2011multimodal}.

To enable the deep model to better capture the correlations between structure and attribute information, multimodal learning method is introduced into the proposed model. The autoencoder-based multimodal learning model \cite{ngiam2011multimodal} is adopted to pre-process the original structure and attribute data. The pre-processing reduces the dimensionality of data from different modalities, specifically removing noise and redundant information to obtain compact high-order features. The correlations across modalities are strengthened between their high-order features. As shown in Figure 2(b), the structure information (adjacency vector) and attribute information (attribute vector) are input separately to a one-layer neural network serving as a pre-processing layer. The use of a pre-training algorithm such as a single-layer RBM enables the pre-processing layer to extract high-order features of each modality. Then, the structure and attribute feature vectors are concatenated and input to the deep autoencoder for further learning. The high-order correlations between structure and attribute will be more facilely learned by deep autoencoder using high-order features obtained by the pre-processing layer. With the subsequent fine-tuning algorithm, the deep autoencoder provides a unified framework to integrate structure and attribute information.

The training goal of the model is preserving the structural and attribute features in the embedding space. The structural features and attribute features are captured by the structural and attribute proximities, respectively. Thus, the model loss function is defined based on the two proximities, as detailed in the next subsection. By fine-tuning the model based on the optimization of the loss function, the obtained representations preserve both the structure and attribute features of the original network.
\begin{figure}[!t]
\centering
\includegraphics{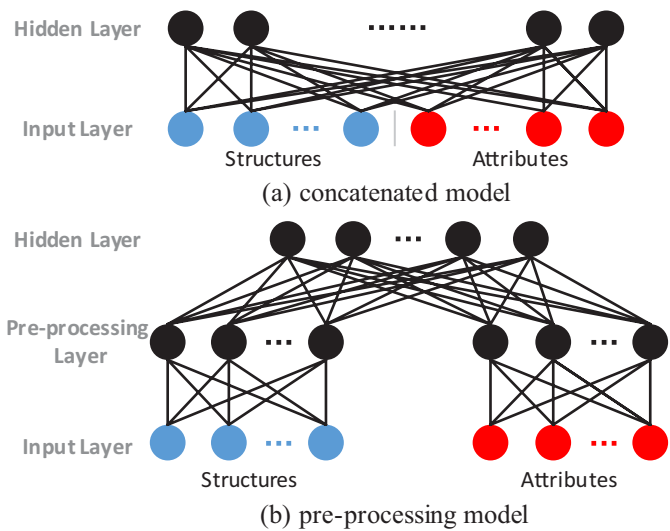}
\caption{Concatenated model and Pre-processing model}
\label{F2}
\end{figure}

In comparison with SDNE \cite{wang2016structural}, which adopts a basic autoencoder to directly reconstruct the input structure information data, the proposed MDNE pre-processes the original adjacency matrix and attribute matrix using a multimodal learning method respectively, and concatenates the resulting high-order structure and attribute features for input to the deep model. The loss function is defined based on the structural and attribute proximities to preserve the structural and attribute features of nodes in the embedding space.

\subsection{Loss Functions}
\label{sec:3.3}
The structural proximity includes the first-order proximity describing the local network structure and the second-order proximity describing the global network structure \cite{wang2016structural}. They are preserved in the loss function to preserve the local and global structural features in low-dimensional embedding space. With the first-order proximity indicating the proximity between directly connected nodes, a corresponding loss function is defined to guarantee that connected nodes with larger weight have a shorter distance in the embedding space, i.e., 
\begin{equation}\label{E1}
{{\cal L}_{1st}} = \sum\limits_{i,j = 1}^n {{s_{ij}}\left\| {{\bf{y}}_i^{(K)} - {\bf{y}}_j^{(K)}} \right\|_2^2}
\end{equation}
. Minimizing ${{\cal L}_{1st}}$ forces the model to preserve the first-order proximity in the embedding space. The second-order proximity represents the similarity of the neighborhood structure between nodes. The neighborhood structure of each node can be described by its adjacency vector. Thus the second-order proximity between two nodes is determined by the similarity of their adjacency vectors, and the goal of the corresponding loss function is to guarantee that nodes with similar adjacency vectors have a short distance in the embedding space. Minimizing the reconstruction error of the input data amounts to maximizing the mutual information between input data and learnt representations \cite{vincent2010stacked}. Intuitively, if the representation allows a good reconstruction of the input data, it means that it has retained much of the information that was present in the input. That is, the MSE-based loss function prompts the basic autoencoder to latently preserve the similarity between input vectors in the embedding space during training. Since the adjacency vector describes the neighborhood structure of each node, minimizing of the reconstruction error of the adjacency vectors preserves the similarity of neighborhood structure (i.e., the second-order proximity) between nodes in the embedding space. Thus, the  loss function based on the second-order proximity is as follows:
\begin{equation}\label{E2}
{{\cal L}_{2nd}} = \sum\limits_{i = 1}^n {\left\| {\left( {{{\bf{\hat s}}_i} - {\bf{s}_i}} \right) \odot \bf{r}_i^s} \right\|_2^2 = \left\| {\left( {\hat S - S} \right) \odot {R^s}} \right\|_F^2} 
\end{equation}
, where $ \odot $ means the Hadamard product, and ${R^s} = \left\{ {\bf{r}_i^s} \right\}_{i = 1}^n$ are the penalty parameters for non-zero adjacency elements. If ${s_{ij}} = 0$, $r_{ij}^s = 1$, else $r_{ij}^s = {\gamma _1} > 1$. Increasing the penalty for the reconstruction error of non-zero elements avoids the reconstruction process's tendency to reconstruct zero elements, making the model robust to sparse networks. Minimizing ${{\cal L}_{1st}}$ and ${{\cal L}_{2nd}}$ imposes a restriction to force the model to preserve the first-order and second-order proximities between nodes.

The attribute proximity of nodes is determined by the similarity of their attribute vectors. The similarity metric of attribute vectors depends on whether the attributes are symmetric or asymmetric. In real-world networks, most of the attributes are highly asymmetric, such as word-counts on citation networks. Moreover, symmetric attributes can also be transformed into asymmetric ones by regarding each ${a_{ij}}$ in node $i$'s attribute vector ${{\bf{a}}_{\bf{i}}}$ as an asymmetric attribute indicating whether node $i$ has attribute value $j$. Therefore, the attribute vectors are treated as highly asymmetric to match real-world circumstances. The asymmetry of both attribute vectors and adjacency vectors results in the same similarity metric of the two data forms. Training the autoencoder to minimize reconstruction error enables the model to preserve the similarity between input vectors in the embedding space \cite{vincent2010stacked}. Meanwhile, experiments in \cite{belkin2003laplacian} shows that minimizing the reconstruction error of the word-count vectors, a kind of highly asymmetric attribute vectors, with a deep autoencoder makes the similar input word-count vectors close to each other in the embedding space. Thus, to preserve the attribute proximity between nodes in the embedding space, the autoencoder is trained to minimize the reconstruction error of the attribute vectors. The corresponding loss function is
\begin{equation}\label{E3}
{{\cal L}_{att}} = \sum\limits_{i = 1}^n {\left\| {\left( {{{\bf{\hat a}}_i} - {\bf{a}_i}} \right) \odot \bf{r}_i^a} \right\|_2^2 = \left\| {\left( {\hat A - A} \right) \odot {R^a}} \right\|_F^2} 
\end{equation}
, where ${R^a} = \left\{ {\bf{r}_i^a} \right\}_{i = 1}^n$ are the penalty parameters for non-zero attribute elements. If ${a_{ij}} = 0$, $r_{ij}^a = 1$, and $r_{ij}^a = {\gamma _2} > 1$ otherwise. The penalty for the reconstruction error of non-zero attribute values reflects that the reconstruction of non-zero elements is more meaningful than the reconstruction of zero ones. This is because there are significantly fewer non-zero elements than zero ones in highly asymmetrical attribute vectors, with non-zero elements much more important in determining the similarity.

The final loss function combines the above structural and attribute proximity loss functions and preserves the structural and attribute proximities between nodes in the embedding space:
\begin{equation}\label{E4}
\begin{array}{rl}
{{\cal L}_{mix}} = &\lambda {{\cal L}_{att}} + \alpha {{\cal L}_{2nd}} + {{\cal L}_{1st}} + \upsilon {{\cal L}_{reg}}\\
 = &\!\lambda \left\| {(\hat A - A) \odot {R^a}} \right\|_F^2 + \alpha \left\| {(\hat S - S) \odot {R^s}} \right\|_F^2\\
 &\!+ \sum\limits_{i,j = 1}^n {{s_{ij}}\left\| {{\bf{y}}_i^{(K)} - {\bf{y}}_j^{(K)}} \right\|_2^2}  + \upsilon {{\cal L}_{reg}}
\end{array}
\end{equation}
, where ${{\cal L}_{reg}}$ is an ${\cal L}2$-norm regularization term to prevent overfitting, and $\lambda $, $\alpha $, and $\upsilon $ are the weight of the attribute proximity loss, second-order proximity loss and regularization term in the loss function. ${{\cal L}_{reg}}$ is defined as:
\begin{equation*}
{{\cal L}_{reg}} = \frac{1}{2}\sum\limits_{k = 1}^K {(\left\| {{W^{(k)}}} \right\|_F^2 + \left\| {{{\hat W}^{(k)}}} \right\|_F^2)}
\end{equation*}
, where ${W^{(k)}},{\hat W^{(k)}},k = 1,\dots,K$ are the weight matrices of the ${k^{th}}$ layer of the encoder and decoder, respectively.

\subsection{Optimization}
\label{sec:3.4}
As presented so far, we seek to minimize the loss function to preserve the structural proximity and attribute proximity in the embedding space. Stochastic gradient descent is a general way to optimize the deep model. However, it is difficult to obtain the optimal result of the model when using stochastic gradient descent directly over randomized weights due to the existence of many local optima \cite{hinton2006reducing}. Otherwise, the gradient descent works well when the initial weights are close to a good solution. Therefore, Deep Belief Network \cite{hinton2006fast} is adopted to pre-train the model and obtain the initial weights, which have been proved to be close to the optimal weights \cite{erhan2010does}. Then, the model is optimized using stochastic gradient descent and the initial weights.

By iterating and updating the parameters until model converges, we obtain the optimal model. Experimental results show that the model optimization converges quickly after the first 10 iterations, and slowly approaches the optimum in the later iterations. Approximately 400 iterations produce the satisfactory results. After proper optimization, informative representations are learned based on the trained model. Algorithm 1 presents the pseudo-code of the proposed method. All the parameters ${W^{(k)}},{\hat W^{(k)}},{\bf{b}^{(k)}},{\bf{\hat b}^{(k)}}$ are signed as $\theta $.
\renewcommand{\algorithmicrequire}{\textbf{Input:}}
\renewcommand{\algorithmicensure}{\textbf{Output:}}
\begin{algorithm}
\caption{MDNE}
\label{A1}
\begin{algorithmic}[1]
\REQUIRE the adjacency matrix $S$, the attribute matrix $A$.
\ENSURE network representation ${Y^{(K)}}$, updated parameters $\theta $.
 \STATE Build pre-processing layer($PPL$), encoder($EC$) and decoder($DC$), pre-train them through Deep Belief Network to obtain the initialized parameters $\theta $;
 \REPEAT
 \STATE ${Y^{(K)}} = EC(PPL(\left[ {S\left. A \right]} \right.),\theta )$, $\left[ {\hat S\left. {\hat A} \right]} \right. = DC({Y^{(K)}},\theta )$;
 \STATE Obtain ${{\cal L}_{mix}}$ based on Eq. (4);
 \STATE Updated parameters $\theta $ through back-propagate algorithm;
 \UNTIL {converge}
 \STATE Obtain the network representations ${Y^{(K)}}$ based on the optimal parameters $\theta $.
 \end{algorithmic}
\end{algorithm}

\subsection{Analysis and Discussions}
\label{sec:3.5}
In this section, we discuss and analyze the proposed model of MDNE.

\textbf{Time Complexity:} 
The time complexity of MDNE is $O((l + f)hr)$, where $l$ is the number of edges, $f$ is the total number of the attributes carried by all the nodes, $h$ is the maximum number of dimensions of the hidden layer, and $r$ is the number of iterations.
Since $h$ and $r$ are independent of the other parameters, the overall training complexity of the model is linear to the sum of the number of edges and attributes carried by all the nodes.

\textbf{New nodes:} A practical issue for network embedding is how to capture evolving networks. Many researches \cite{huang2017accelerated,Liao2017Attributed} have shown interest in dealing with dynamic topological structures and node attributes. Since newly arriving nodes are an important factor for evolving networks, the proposed method provides a possible way to represent them. If new nodes have observable links connecting to existing nodes and bringing attribute information as well, their representations can be obtained by feeding their adjacency vectors and attribute vectors into the finely trained model. If the new nodes lack structure or attribute information, most existing methods cannot handle them \cite{wang2016structural}. However, MDNE can learn the representations of the new nodes lacking one modality of information by replacing the missing vectors with zero vectors and inputting the existing vectors together with the zero vectors to the trained model.

\section{Experimental Results}
\label{sec:4}
In this section, we empirically evaluate the effectiveness and generality of the proposed algorithm. First, the experimental setup is introduced, including datasets, baseline methods and parameter settings. We also investigate the convergence of MDNE, and verify the ability of all methods to reconstruct the network structure. Then, the comparisons of the proposed method and baselines are conducted on three real-world network analysis tasks, i.e., link prediction, attribute prediction and classification, to verify the ability of the obtained representations. Finally, the parameter sensitivity and the impact of pre-processing are discussed. Experiments run on a Dell Precision Tower 5810 with an Intel Xeon CPU E5-1620 v3 at 3.50 GHz and 16 GB of RAM.

\subsection{Experiment Setup}
\label{sec:4.1}
\subsubsection{Datasets}
Four real-world network datasets are used in this work, including two citation networks and two social networks. Considering the characteristics of these datasets, one or more datasets are chosen to evaluate the performances on each network analysis task. Four datasets are described as follows.

\textbf{cora:} cora\footnote{http://linqs.cs.umd.edu/projects//projects/lbc/index.html} is a citation network which contains 2,708 nodes and 5,278 edges. Each node indicates a machine learning paper, and the edge indicates the citation relation between papers. After stemming and removing stop-words, a vocabulary of 1433 unique words is regarded as the attribute information of papers. Each attribute indicates the absence/presence of the corresponding word in papers. These papers are classified into one of the following seven classes: Case Based, Genetic Algorithms, Neural Networks, Probabilistic Methods, Reinforcement Learning, Rule Learning, and Theory.

\textbf{citesee:} citeseer is a citation network which consists of 3,312 nodes and 4,551 edges. Similarly, nodes and edges represent scientific publications and their citations, respectively. The vocabulary of size 3,703 words is extracted and set as the attributes. These papers are classified into one of the following six classes: Agents, AI, DB, IR, ML, HCI.

\textbf{UNC, Oklahoma:} They are two Facebook sub-networks, which respectively contains 18,163 students from the University of North Carolina and 17,425 students from University of Oklahoma, and also with their seven anonymized attributes: status, gender, major, second major, dorm/house, high school, class year. Note that not all of the students have the seven attributes available.
\begin{table}[!t]
\renewcommand{\arraystretch}{1.3}
\caption{Dataset statistics}
\label{T2}
\centering
\begin{tabular}{|c|c|c|c|}
\hline
Dataset & \# nodes & \# edges & \# attributes \\
\hline
UNC &	 18163 &	766800 &	2788 \\
\hline
Oklahoma &	17425 &	892528 &	2305 \\
\hline
citeseer &	3312 &	4551 &	3703 \\
\hline
cora &	 2708 &	5278 &	 1433 \\
\hline
\end{tabular}
\end{table}

The statistics of the four datasets are summarized in Table 2. Experiments are conducted on both weighted and unweighted, small and large networks. Diverse datasets allow us to evaluate whether the proposed network embedding method has a better performance on networks with different characteristics.

\subsubsection{Baseline Methods}
Five typical methods are chosen to be baselines.

\textbf{LE \cite{belkin2002laplacian}:} It provides Laplacian Eigenmaps and spectral techniques to embed the data into a latent low-dimensional space. The solution reflects the features of the network structure.

\textbf{node2vec \cite{grover2016node2vec}:} It samples the network structure by the biased random walk. By regarding the paths as sentences, it adopts the natural language processing model to generate network embedding. The hyper-parameters $p$ and $q$ introduce breadth-first sampling and depth-first sampling in the random walk. It can recover DeepWalk when $p$ and $q$ are set to 1.

\textbf{SDNE \cite{wang2016structural}:} It exploits the first-order proximity and second-order proximity to preserve the local and global network structure. A deep model is adopted to address the highly non-linear structure and sparsity problem of networks.

\textbf{AANE \cite{huang2017accelerated}:} It proposes a scalable and efficient framework which incorporates node attribute proximity into network embedding. It processes each node efficiently by decomposing the complex modeling and optimization into many sub-problems.

\textbf{ASNE \cite{Liao2017Attributed}:} It adopts a multilayer neural network to capture the complex interactions between features which denote the ID and attributes of nodes, and the proposed framework performs network embedding by preserving the structural proximity and attribute proximity of nodes in the paths generated by the random walk.

The first three methods are pure structure-based methods, and the others integrate attribute and structure information into network embedding.

\subsubsection{Parameter Settings}
The depth of neural networks and the number of neurons are essential factors in learning effect. Recent evidences \cite{wang2016structural,simonyan2014very,szegedy2015going} reveal that the number of stacked layers (depth) and neurons should be neither too large nor too small. Large numbers of layers and neurons increase the difficulty of training the model, and bring over-fitting problem. However, too few layers and neurons fail to extract effective low-dimensional representations \cite{zeiler2014visualizing}, especially for large-scale datasets. Therefore, we vary MDNE's neural network structure according to different datasets, as shown in Table 3. Two numbers in the first layer and second layer indicate the dimensions of the vectors related to the structure and attribute data, respectively.

We implemented of MDNE using TensorFlow\footnote{https://www.tensorflow.org/}. We fine-tuned the loss function hyper-parameters $\lambda ,\alpha ,\upsilon ,{\gamma _1},{\gamma _2}$ using grid search based on the performance of the network reconstruction \cite{wang2016structural}, which is introduced as a basic quality criterion of the proposed method in Section 4.3. We first perform a parameter sweep setting $\lambda ,\alpha ,\upsilon ,{\gamma _1},{\gamma _2} = \{0, 0.01, 0.1, 1, 10, 100, 1000\}$ on each dataset. They are tuned one by one iteratively until all of them are converged. Then every hyper-parameter is further fine-tuned by grid search on a smaller space around optimal value got in previous search for each dataset.
\begin{table}[!t]
\renewcommand{\arraystretch}{1.3}
\caption{Neural Network Structures}
\label{T3}
\centering
\begin{tabular}{|c|c|}
\hline
Dataset & \# nodes in each layer \\
\hline
cora &	 (2708,1433)-(300,200)-128 \\
\hline
citeseer &	(3312,3703)-(250,250)-128 \\
\hline
UNC &	(18163,2788)-(3000,500)-128 \\
\hline
Oklahoma &	(17425,2305)-(3600,650)-128 \\
\hline
\end{tabular}
\end{table}

The parameters of the baseline methods are adjusted to the optimal values as given in their researches. For the sake of fairness, we set the embedding dimensions of all the methods $d=128$ on different tasks.

\subsection{Convergence}
Experiments are conducted to investigate the convergence property of MDNE. We vary the number of iterations from 0 to 800 and plot the corresponding value of loss function on a citation network cora and a social network UNC. The learning curves are shown in Figure 3. The result indicates that MDNE convergences at about 400 iterations on different datasets. Although the performance may be better with more iterations, 400 iterations have achieved the best result among baselines. To balance the effectiveness and efficiency of MDNE, the model is trained about 400 iterations in experiments.

\begin{figure}[!t]
\centering
\includegraphics{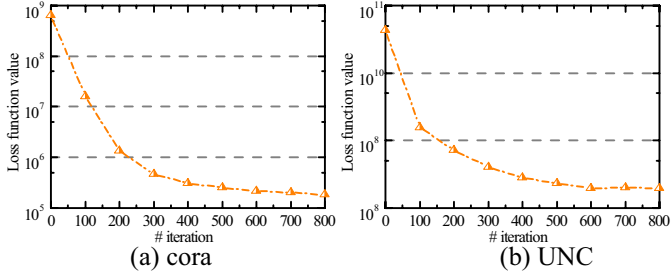}
\caption{Convergence of MDNE on cora and UNC datasets.}
\label{F3}
\end{figure}

\subsection{Network Reconstruction}
Network reconstruction verifies the ability of the method to reconstruct the network structure, which is also a basic requirement for network embedding methods. Given the learned network representations, all links in the original network need to be predicted. The way to predict the links is ranking all node pairs based on their similarity and predicting that a certain number of top pairs are linked by edges. The cosine distance of learned vectors measures the similarities between nodes. The higher-ranking node pairs are more likely to have links in the original network. The evaluation indicator is $precision@k$ \cite{wang2016structural}, referring to the ratio of the top $k$ node pairs to be connected in the original network. A larger $precision@k$ indicates the better performance of the reconstruction.

\begin{figure}[!t]
\centering
\includegraphics{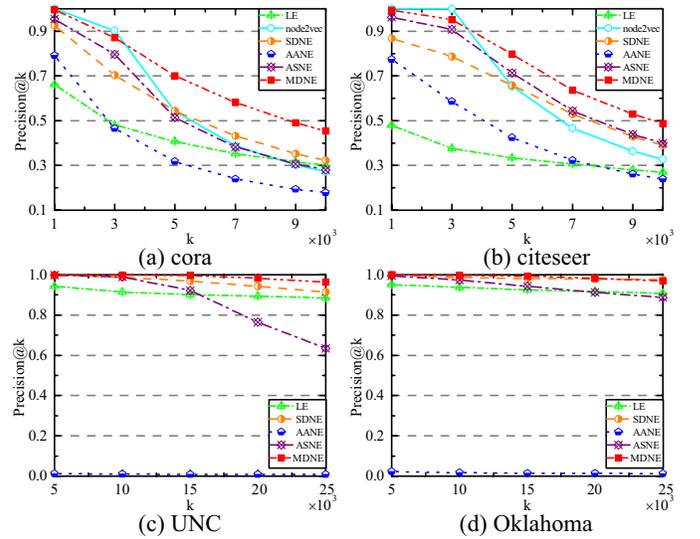}
\caption{Network Reconstruction performance on different datasets. Node2vec can't obtain results on UNC and Oklahoma which have more than 10,000 nodes, due to an out of memory problem. $k$ is set based on the network scale.}
\label{F4}
\end{figure}

\begin{table}[!t]
\renewcommand{\arraystretch}{1.3}
\caption{$precision@k$ of Network Reconstruction on cora and citeseer datasets}
\label{T4}
\centering
\tiny
\begin{tabular}{|c|c|c|c|c|c|c|c|}
\hline
     &Algorithm & $P$@1000 & $P$@3000 & $P$@5000 & $P$@7000 & $P$@9000 & $P$@10000 \\
\hline
               & LE &  0.661          & 0.481          & 0.408           & 0.353           & 0.316           & 0.300 \\
  & node2vec & \textbf{1.000} & \textbf{0.903}& 0.542           & 0.388           & 0.302           & 0.272 \\
cora  & SDNE &  0.924          & 0.703          & 0.543           & 0.432           & 0.353           & 0.323 \\
       & AANE &  0.792          & 0.465          & 0.318           & 0.239           & 0.194           & 0.179 \\
       & ASNE  &  0.954          & 0.796          & 0.514           & 0.383           & 0.307           & 0.281 \\
      & MDNE & 0.996           & 0.871          & \textbf{0.701} & \textbf{0.581} & \textbf{0.491} & \textbf{0.455} \\
\hline
              & LE &  0.480          & 0.376          & 0.334           & 0.307           & 0.280           & 0.269 \\
  & node2vec & \textbf{1.000}& \textbf{1.000} & 0.654           & 0.467           & 0.364           & 0.327 \\
citeseer  & SDNE &  0.869  & 0.787          & 0.658           & 0.530           & 0.430           & 0.390 \\
       & AANE &  0.774          & 0.586          & 0.424           & 0.323           & 0.262           & 0.239 \\
       & ASNE  &  0.962          & 0.908          & 0.713           & 0.543           & 0.438           & 0.400 \\
      & MDNE & 0.994           & 0.951          & \textbf{0.798} & \textbf{0.637} & \textbf{0.530} & \textbf{0.488} \\
\hline
\end{tabular}
\end{table}

\begin{table}[!t]
\renewcommand{\arraystretch}{1.3}
\caption{$precision@k$ of Network Reconstruction on UNC and Oklahoma datasets}
\label{T5}
\centering
\tiny
\begin{tabular}{|c|c|c|c|c|c|c|}
\hline
     &Algorithm & $P$@5000 & $P$@10000 & $P$@15000 & $P$@20000 & $P$@25000 \\
\hline
               & LE &  0.942          & 0.915            & 0.901           & 0.894           & 0.885  \\
UNC & SDNE &  0.997          & 0.988          & 0.968           & 0.943           & 0.915  \\
       & AANE &  0.012          & 0.010          & 0.008           & 0.008           & 0.008  \\
       & ASNE  &  \textbf{0.999} & 0.989          & 0.922           & 0.765           & 0.635  \\
      & MDNE & 0.998 & \textbf{0.998} & \textbf{0.997} & \textbf{0.982} & \textbf{0.963} \\
\hline
              & LE &  0.952          & 0.938          & 0.925           & 0.916           & 0.907   \\
Oklahoma & SDNE &  0.998  & 0.986          & 0.981           & 0.978           & 0.976  \\
       & AANE &  0.022          & 0.018          & 0.015           & 0.014           & 0.013    \\
       & ASNE  &  0.995          & 0.974          & 0.943           & 0.914           & 0.888    \\
      & MDNE & \textbf{0.999}& \textbf{0.996} & \textbf{0.993} & \textbf{0.983} & \textbf{0.969} \\
\hline
\end{tabular}
\end{table}

Network reconstruction has been performed on the four datasets, the results of which are shown in Figure 4. Also, Table 4 and Table 5 provide the numeric results helping to compare the close curves. Numbers in bold represent the best result in each column. Compared to UNC and Oklahoma, the performance of all the methods visibly decrease on cora and citeseer. This is because cora and citeseer have sparsity problem, as their average degree is much smaller than that of UNC and Oklahoma.

LE, which is a shallow model-based method, has poor performance. It indicates that going deep enhance the model's generalization ability, and helps to capture the high non-linearity of network structures. SDNE adopts deep autoencoder model but only uses structure information. Its inferior performance demonstrates the usefulness of attribute information in learning better node representations. Node2vec is slightly better than MDNE on cora and citeseer network when $k=1000$ to $k=3000$. The reason might be that node2vec can capture the higher-order proximity between nodes by random walks in the network.

AANE has relatively poor performance, especially on UNC and Oklahoma. This is because AANE only considers the first-order proximity, and its performance largely depends on the computation of attribute similarity under full attribute space, while attribute similarity of nodes computed under high-dimensional attribute space explicitly on certain networks has little discriminability. ASNE has slightly inferior performance because it is hard to capture the non-linear correlations of structure and attribute information, as it pre-processes structure and attribute data linearly before concatenating them. MDNE has the best performance on four datasets in most cases. The good performance of MDNE is because it adopts a deep model to learn non-linear features, and uses multimodal learning method to better capture the correlations of attribute and structure, and preserves the attribute proximity by minimizing the reconstruction error instead of computing the attribute similarity explicitly.

\subsection{Link Prediction and Attribute Prediction}
In this section, we evaluate the ability of the learned representations to predict missing links and attributes in the network, which is a practical task in real-world applications.

\subsubsection{Link Prediction}
Link prediction is the prediction of missing links based on the existing information. After hiding $5\% \sim 45\%$ of links randomly, the left network is utilized as a sub-dataset to perform network embedding. The test set consisted of positive instances and negative instances. The hidden links are taken as positive instances and the same ratio of unconnected node pairs in the original network are randomly selected to be negative instances. Similarities between the learned representations of nodes in the test set are calculated and are sorted in descending order. A higher ranking of a node pair corresponds to a greater possibility for them to be connected. Area Under the ROC Curve (AUC) is adopted as the evaluation metric as it is commonly used to measure the quality of classification based on ranking. A large AUC indicates good performance. If an algorithm ranks all positive instances higher than all negative instances, the AUC is 1. The above steps are repeated 10 times and the average AUC is taken as the final result. All methods had extremely poor performance on the cora and citeseer networks, as the low average degrees of the two networks make link prediction very hard. Thus we only show the results on the UNC and Oklahoma networks in Figure 5 and Table 6. Numbers in bold represent the highest performance in each column.

\begin{figure}[!t]
\centering
\includegraphics{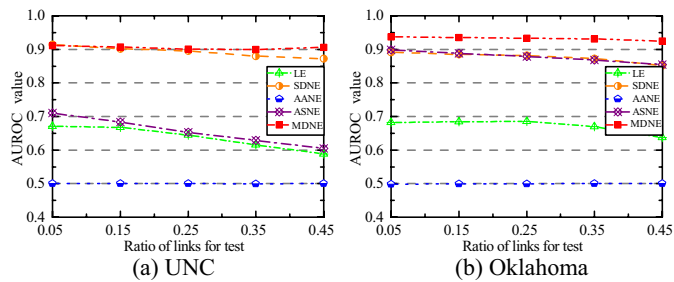}
\caption{Link prediction performance on UNC and Oklahoma datasets.}
\label{F5}
\end{figure}

\begin{table}[!t]
\renewcommand{\arraystretch}{1.3}
\caption{AUC of Link Prediction on UNC and Oklahoma datasets}
\label{T6}
\centering
\begin{tabular}{|c|c|c|c|c|c|c|}
\hline
     &Test Ratio & 0.05    & 0.15    & 0.25    & 0.35     & 0.45    \\
\hline
               & LE & 0.670 & 0.668  & 0.644  & 0.616   & 0.588   \\
UNC & SDNE & \textbf{0.915} & 0.902   & 0.896  & 0.880  & 0.873  \\
       & AANE &  0.501  & 0.500  & 0.500  & 0.500   & 0.499  \\
       & ASNE  &  0.711  & 0.683  & 0.653  & 0.629   & 0.605  \\
      & MDNE & 0.912  & \textbf{0.907} & \textbf{0.901} & \textbf{0.900} & \textbf{0.906} \\
\hline
               & LE & 0.682 & 0.685  & 0.685  & 0.670   & 0.637   \\
Oklahoma & SDNE &0.891 & 0.885   & 0.882  & 0.873  & 0.852  \\
       & AANE &  0.498  & 0.499  & 0.500  & 0.501   & 0.500  \\
       & ASNE  &  0.899  & 0.889  & 0.879  & 0.868   & 0.855  \\
      & MDNE & 0.937  & \textbf{0.935} & \textbf{0.933} & \textbf{0.931} & \textbf{0.924} \\
\hline
\end{tabular}
\end{table}

\begin{table}[!t]
\renewcommand{\arraystretch}{1.3}
\caption{$p$-value of Friedman Test on MDNE with baselines for Link Prediction}
\label{T7}
\centering
\begin{tabular}{|c|c|c|}
\hline
Baseline & UNC & Oklahoma \\
\hline
 LE      & 1.125$e-5$ & 1.125$e-5$  \\
\hline
SDNE & 0.0084 & 1.125$e-5$ \\
\hline
AANE & 1.125$e-5$ & 1.125$e-5$  \\
\hline
ASNE &  1.125$e-5$ & 1.125$e-5$  \\
\hline
\end{tabular}
\end{table}

Compared with the shallow model-based methods LE and AANE, the deep model-based methods MDNE, SDNE and ASNE perform significantly better. This is because the deep model can better capture highly non-linear network structures. The reason for the extremely poor performance of AANE is similar to that on network reconstruction task. SDNE and MDNE have good performance since preserve both the first-order and second-order proximities between nodes in the embedding space. MDNE is slightly better than SDNE which does not preserve attribute features in the learned representations. This result justifies the usefulness of attribute information in link prediction.

Friedman test is conducted to better endorse the superiority of MDNE with respect to other methods. The $p$-values are computed based on the ranking for the AUC value of MDNE with each method on different sub-datasets with different test ratios. In Table 7, all the $p$-values are less than 0.05. The results show that the performance of MDNE is significantly different from the compared methods on link prediction task. The $p$-value of MDNE with SDNE on UNC is slightly higher than others because SDNE is slightly better than MDNE when the ratio of links for test is 5\%.

The network becomes sparse with the ratio of links for test increasing, and the AUC of MDNE is stable while that of other methods dropped. It indicates that the penalty for non-zero elements in the loss function improves MDNE’s performance in dealing with sparse networks. Such an advantage is pivotal for downstream applications since links are often sparse, especially in large-scale real-world networks. Despite the link prediction task's favoring pure structure-based methods, our MDNE outperforms the others. This demonstrates the effectiveness of the learned representations in predicting missing links.

\subsubsection{Attribute Prediction} 
Attribute prediction refers to predicting unknown attribute values of nodes based on the obtained information. It has enjoyed increasing interest in network analysis tasks. For example, in social network recommendation, predicting attribute features is essential to help users to locate their interested information \cite{cai2018three}.

In attribute prediction experiments, $5\%\sim45\%$ of the attribute values (including value 1 and value 0) in the original network are hidden randomly, i.e., $5\%\sim45\%$ of ${a_{jk}}$ in the attribute matrix $A$ are hidden. They are set as the test set. The left attribute information and structure information is trained to learn the representations of nodes. The obtained representations are used to predict the attributes in the test set. Assuming that the attribute $k$ of node $j$ is hidden, here is the way to predict. Similarities between node $j$ with all the other nodes in the embedding space are calculated, denoting as $sim_{ij},i = 1,\dots,n$. We denote the top 10 nodes with the highest similarity to $j$ as set $N_S$. ${N_p} = \left\{ {i|{a_{ik}} = 1,i \in {N_S}} \right\}$, is the set of nodes with the attribute value $k=1$ in the above set. Similarly, ${N_n} = \left\{ {i|{a_{ik}} = 0,i \in {N_S}} \right\}$, is the set of the rest of nodes with the attribute value $k=0$ in $N_S$. $p = {{\sum\limits_{i \in {N_p}} {si{m_{ij}}} } / {\sum\limits_{i \in {N_n}} {si{m_{ij}}} }}$ is calculated, which indicates the possibility of that the ${a_{jk}}$, i.e., attribute $k$ of node $j$, is 1. All ${a_{jk}}$ in the test set are sorted in descending order of $p$. AUC is the metric to evaluate the $p$ ranking list. A high AUC value indicates the high accuracy of the prediction. The result shows as in Figure 6.

\begin{figure}[!t]
\centering
\includegraphics{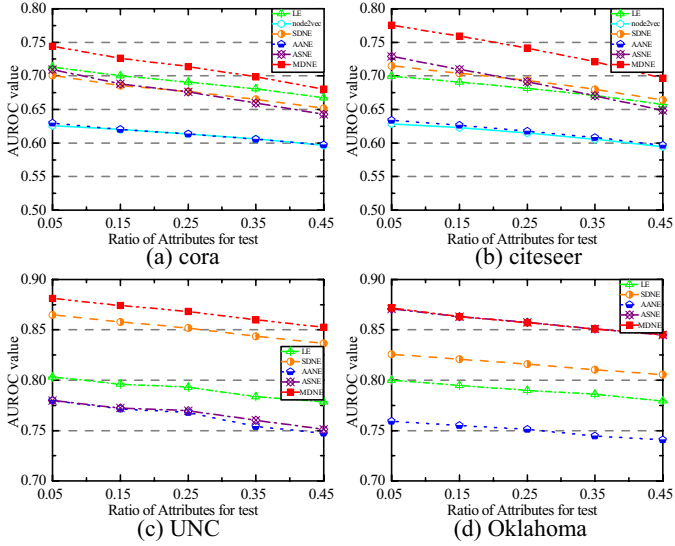}
\caption{Attribute prediction performance on different datasets.}
\label{F6}
\end{figure}

\begin{table}[!t]
\renewcommand{\arraystretch}{1.3}
\caption{$p$-value of Friedman Test on MDNE with baselines for Attribute Prediction}
\label{T8}
\centering
\begin{tabular}{|c|c|c|c|c|}
\hline
Baseline & cora & citeseer & UNC & Oklahoma \\
\hline
 LE      & 1.125$e-5$ & 1.125$e-5$ & 1.125$e-5$ & 1.125$e-5$  \\
\hline
SDNE & 1.125$e-5$ & 1.125$e-5$  & 1.125$e-5$ & 1.125$e-5$\\
\hline
AANE & 1.125$e-5$ & 1.125$e-5$ & 1.125$e-5$ & 1.125$e-5$  \\
\hline
ASNE &  1.125$e-5$ & 1.125$e-5$ &    /    &   /    \\
\hline
\end{tabular}
\end{table}

The performances of LE, node2vec and SDNE are worse than that of MDNE. The reason is their lacking of consideration of preserving the attribute features in the embedding space, which is important for predicting missing attributes of nodes. AANE still has poor performance. It is because the attribute affinity matrix adopted by AANE is calculated based on the full attribute space, which decreases the discriminability of the representations. Compared with ASNE, the superior performance of our MDNE credits the pre-processing of the original attribute and structure information based on multimodal learning method. The high-order features of the attribute vectors and adjacency vectors obtained by the pre-processing layer help the successive layers better extract the high-order correlations between the structure and attribute feature of nodes.

Also, the Friedman test is conducted on MDNE with others. The $p$-values are listed in Table 8, all of which are less than 0.05. The results show that the performance of MDNE is significantly different from baselines on attribute prediction task.

The attribute sparsity of different datasets is quite different, as the average number of attributes of each node is 34.3, 31.7, 5.4 and 5.3 on cora, citeseer, UNC and Oklahoma respectively. Moreover, the attributes of each network become sparse with the ratio of the test set increasing. The proposed method has good performance in all cases. This demonstrates that MDNE is effective in attribute prediction tasks and is robust to networks with different extent of attribute sparseness.

\subsection{Classification}

Classification is one of the important tasks in network analysis. It classifies nodes based on their features. The representations generated are used as features. The widely used classifier LIBLINEAR \cite{fan2008liblinear} is adopted. A portion of node representations and their labels are taken as the training set, and the rest to be the test set. For a fair comparison, the test ratio varies from $10\%\sim90\%$ by an increment of $10\%$. F-measure is a commonly adopted metric for binary classification. Micro-F1 \& Macro-F1 are employed to judge the classification quality. Macro-average is defined as an arithmetic average of F-measure of all the label categories, and Micro-average is the harmonic mean of average precision and average recall. For both metrics, the higher values indicate better performance. For each training ratio, we randomly split the training set and the test set for 10 times and report the average result as Figure 7. The experiment is conducted on citeseer and cora, since they are the only datasets containing class labels for nodes.

\begin{figure}[!t]
\centering
\includegraphics{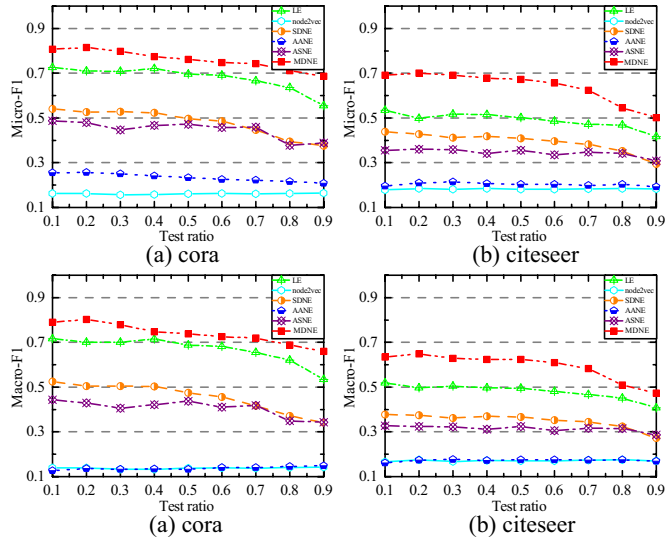}
\caption{Classification performance on cora and citeseer datasets.}
\label{F7}
\end{figure}

MDNE always has the best performance in all cases. Although node2vec has satisfactory performance on network reconstruction task, it returns the disappointing result on classification task. This shows the representations learned by node2vec have task preference. AANE still has poor performance. We replay the classification experiments with the non-linear kernel SVM classifier and 5-fold cross validation. The performance of AANE is improved. This is because AANE is hard to capture the non-linear correlations between structure and attribute features, and the representations learned by AANE are non-linear. The SVM with the non-linear kernel has the classification ability with non-linear representations, which is difficult for the linear classifier LIBLINEAR to deal with. MDNE has good performance on both LIBLINEAR and SVM. Considering LIBLINEAR has advantages in time complexity, it is beneficial that the learned representations are suitable for linear classifiers. The poor performance of ASNE is due to its lacking of non-linear pre-processing of the original structure and attribute information. The non-linear pre-processing of the adjacency vector and attribute vector can help the model to capture the high-order correlations between the two information in the subsequent learning. SDNE and LE are worse than MDNE, as they do not consider attribute information when embedding networks. The significant improvement of MDNE over baselines proves that adopting multimodal deep model and optimizing the loss function defined based on the structural proximity and attribute proximity are able to learn effective representations for classification tasks.

\subsection{Parameters Sensitivity and the Impact of Pre-Processing}
In this section, we investigate how different choices of $\lambda$ and embedding dimensions, along with the consideration of pre-processing affect the performance of MDNE on the cora dataset. The results of classification tasks with different test ratios are reported. The results from other tasks on other datasets are omitted as they are similar.
\begin{figure}[!t]
\centering
\includegraphics{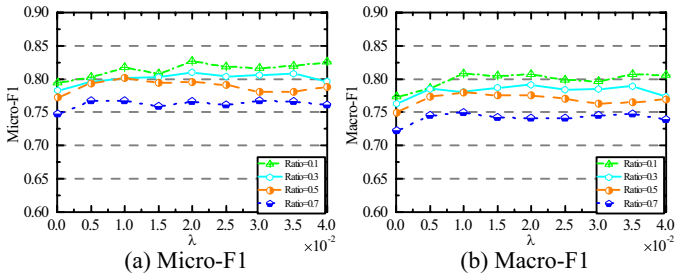}
\caption{Classification performance of MDNE on cora dataset with different weight of the attribute proximity loss $\lambda$.}
\end{figure}

\subsubsection{The weight of the attribute proximity loss $\lambda$}
The hyper-parameter $\lambda$ adjusts the importance of attribute proximity loss in the loss function. The weight of structural proximity loss is fixed as $\alpha  = 0.5$. Then the $\lambda$ will determine the relative importance between the attribute proximity loss and the structural proximity loss.

Figure 8 shows the impact of $\lambda$ on the range of [0, 0.04] at an interval of 0.005. The slightly improving performance on $\lambda  = [0, 0.02]$ shows that attribute proximity loss plays an important role in learning network representations. The performance relatively stabilized on $\lambda  = [0.02, 0.04]$ indicates that the performance of MDNE is not sensitive to values on this range, which means the value of $\lambda$ is suitable for the proposed model in a wide range in real-world applications. The great difference between $\lambda$ and the weight of structural proximity  loss is due to the inherent characteristics of dataset cora. The total number of edges is 5278, and the total number of attribute values is 49216. That is, the attribute proximity loss ${{\cal L}_{att}}$ is much larger than the structural proximity loss. To balance the different effect from them, the smaller weight of ${{\cal L}_{att}}$ is necessary. Besides, compared with Figure 7, when $\lambda  = 0$, which means the attribute proximity loss is ignored in loss function, MDNE still outperforms baselines. The observation indicates that the structure of the proposed multimodal deep autoencoder with the pre-training algorithm is able to capture the highly non-linear relationship between structure and attribute features even without attribute proximity loss in the loss function.

\begin{figure}[!t]
\centering
\includegraphics{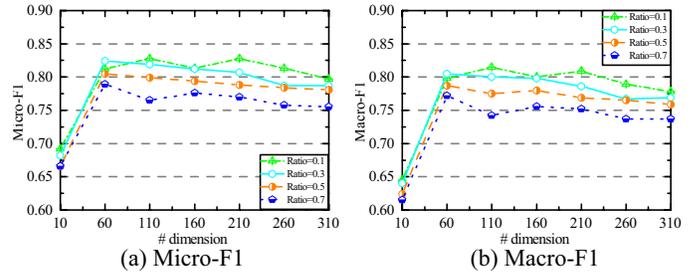}
\caption{Classification performance of MDNE on cora dataset with different embedding dimensions.}
\label{F9}
\end{figure}

\subsubsection{Embedding dimensions}
The effect of the embedding dimensions on classification performance is shown in Figure 9. The performance gets better as the number of dimensions increasing initially. When the number of dimensions is larger than a threshold, the performance becomes stable. The reason is twofold. When the number of dimensions is small, more useful information is incorporated into representations with the number of dimensions increasing and the performance also increase. However, the too large number of dimensions also bring noise and redundant information which weaken the classification ability of the representations. Thus, it is important to select a reasonable embedding dimension. It is observed from Figure 9 that the proposed method is not very sensitive to embedding dimensions when the number of dimensions is larger than 60. Taking into account the accuracy and complexity of nodes, the embedding dimensions of MDNE is set as 128 in our experiments.
\begin{figure}[!t]
\centering
\includegraphics{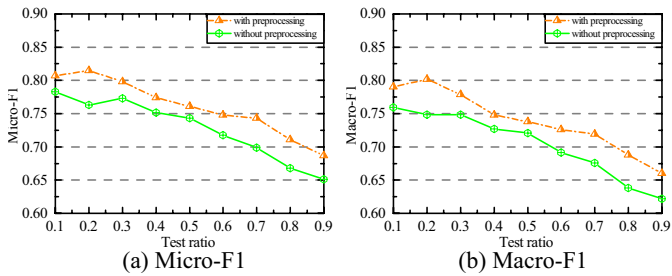}
\caption{Classification performance of MDNE on cora dataset with and without preprocessing.}
\label{F10}
\end{figure}

\subsubsection{Pre-processing}
Figure 10 shows the results of MDNE with and without the pre-processing procedure. The model with the pre-processing procedure, which corresponds to Figure 2(b), has the structure of \{(2708,1433)-(300,200)-128\}. Except for the pre-processing layer, the subsequent deep model has an input layer with the concatenated high-order features and an output layer. The model without the pre-processing procedure, which corresponds to Figure 2(a), has the structure of \{(2708,1433)-500-128\}. The corresponding deep model has an input layer with concatenated original vectors, a hidden layer, and an output layer. The total number of weight parameters in the model without the pre-processing procedure is larger than that in the model with the pre-processing procedure. As Figure 10 shows, although the pre-processing model has smaller computation complexity, its result is slightly better. Moreover, compared with Figure 7, the proposed method without the pre-processing procedure is still better than baselines. It demonstrates that besides the pre-processing procedure, both the deep model and loss function of the proposed method contribute to the good performance of MDNE.

\section{Conclusion}
In this paper, a Multimodal Deep Network Embedding method is proposed for learning informative network representations by integrating the structure and attribute information of nodes. Specifically, the deep model comprising of multiple layers of non-linear functions is adopted to capture the non-linear network structure and the complex interactions with node attributes. In order to better extract the high-order correlations between the topological structures and attributes of nodes, the multimodal learning method is adopted to pre-process the original structure and attribute data. The structural proximity and attribute proximity are utilized to describe the structure and attribute features of the network, respectively. The model loss function is defined based on the two proximities. Minimizing the loss function preserves both proximities in the embedding space. Experiments are conducted on four real-world networks to evaluate the performance of the representations obtained. Compared with baselines, the result demonstrates that MDNE offers superior performance on various real-world applications. In the future, we will consider improving the efficiency of MDNE through a parallel processing framework and expanding the model to learn task-oriented representations combined with the requirements from specific applications.


%

%

\section*{Acknowledgments}
The authors would like to thank the anonymous referees for their critical appraisals and useful suggestions.

\ifCLASSOPTIONcaptionsoff
  \newpage
\fi



\bibliographystyle{IEEEtran}
\bibliography{Manuscript.bbl}

\begin{thebibliography}{10}
\providecommand{\url}[1]{#1}
\csname url@samestyle\endcsname
\providecommand{\newblock}{\relax}
\providecommand{\bibinfo}[2]{#2}
\providecommand{\BIBentrySTDinterwordspacing}{\spaceskip=0pt\relax}
\providecommand{\BIBentryALTinterwordstretchfactor}{4}
\providecommand{\BIBentryALTinterwordspacing}{\spaceskip=\fontdimen2\font plus
\BIBentryALTinterwordstretchfactor\fontdimen3\font minus
  \fontdimen4\font\relax}
\providecommand{\BIBforeignlanguage}[2]{{%
\expandafter\ifx\csname l@#1\endcsname\relax
\typeout{** WARNING: IEEEtran.bst: No hyphenation pattern has been}%
\typeout{** loaded for the language `#1'. Using the pattern for}%
\typeout{** the default language instead.}%
\else
\language=\csname l@#1\endcsname
\fi
#2}}
\providecommand{\BIBdecl}{\relax}
\BIBdecl

\bibitem{perozzi2014deepwalk}
B.~Perozzi, R.~Al-Rfou, and S.~Skiena, ``Deepwalk: Online learning of social
  representations,'' in \emph{Proceedings of the 20th ACM SIGKDD international
  conference on Knowledge discovery and data mining}.\hskip 1em plus 0.5em
  minus 0.4em\relax ACM, 2014, pp. 701--710.

\bibitem{grover2016node2vec}
A.~Grover and J.~Leskovec, ``node2vec: Scalable feature learning for
  networks,'' in \emph{Proceedings of the 22nd ACM SIGKDD international
  conference on Knowledge discovery and data mining}.\hskip 1em plus 0.5em
  minus 0.4em\relax ACM, 2016, pp. 855--864.

\bibitem{cao2015grarep}
S.~Cao, W.~Lu, and Q.~Xu, ``Grarep: Learning graph representations with global
  structural information,'' in \emph{Proceedings of the 24th ACM International
  on Conference on Information and Knowledge Management}.\hskip 1em plus 0.5em
  minus 0.4em\relax ACM, 2015, pp. 891--900.

\bibitem{hu2013exploiting}
X.~Hu, L.~Tang, J.~Tang, and H.~Liu, ``Exploiting social relations for
  sentiment analysis in microblogging,'' in \emph{Proceedings of the sixth ACM
  international conference on Web search and data mining}.\hskip 1em plus 0.5em
  minus 0.4em\relax ACM, 2013, pp. 537--546.

\bibitem{tang2013exploiting}
J.~Tang, H.~Gao, X.~Hu, and H.~Liu, ``Exploiting homophily effect for trust
  prediction,'' in \emph{Proceedings of the sixth ACM international conference
  on Web search and data mining}.\hskip 1em plus 0.5em minus 0.4em\relax ACM,
  2013, pp. 53--62.

\bibitem{mcpherson2001birds}
M.~McPherson, L.~Smith-Lovin, and J.~M. Cook, ``Birds of a feather: Homophily
  in social networks,'' \emph{Annual review of sociology}, vol.~27, no.~1, pp.
  415--444, 2001.

\bibitem{marsden1988homogeneity}
P.~V. Marsden, ``Homogeneity in confiding relations,'' \emph{Social networks},
  vol.~10, no.~1, pp. 57--76, 1988.

\bibitem{luo2011cauchy}
D.~Luo, F.~Nie, H.~Huang, and C.~H. Ding, ``Cauchy graph embedding,'' in
  \emph{Proceedings of the 28th International Conference on Machine Learning
  (ICML-11)}, 2011, pp. 553--560.

\bibitem{cui2017survey}
P.~Cui, X.~Wang, J.~Pei, and W.~Zhu, ``A survey on network embedding,''
  \emph{arXiv preprint arXiv:1711.08752}, 2017.

\bibitem{bengio2009learning}
Y.~Bengio \emph{et~al.}, ``Learning deep architectures for ai,''
  \emph{Foundations and trends{\textregistered} in Machine Learning}, vol.~2,
  no.~1, pp. 1--127, 2009.

\bibitem{hinton2006reducing}
G.~E. Hinton and R.~R. Salakhutdinov, ``Reducing the dimensionality of data
  with neural networks,'' \emph{science}, vol. 313, no. 5786, pp. 504--507,
  2006.

\bibitem{ngiam2011multimodal}
J.~Ngiam, A.~Khosla, M.~Kim, J.~Nam, H.~Lee, and A.~Y. Ng, ``Multimodal deep
  learning,'' in \emph{Proceedings of the 28th international conference on
  machine learning (ICML-11)}, 2011, pp. 689--696.

\bibitem{wang2016structural}
D.~Wang, P.~Cui, and W.~Zhu, ``Structural deep network embedding,'' in
  \emph{Proceedings of the 22nd ACM SIGKDD international conference on
  Knowledge discovery and data mining}.\hskip 1em plus 0.5em minus 0.4em\relax
  ACM, 2016, pp. 1225--1234.

\bibitem{roweis2000nonlinear}
S.~T. Roweis and L.~K. Saul, ``Nonlinear dimensionality reduction by locally
  linear embedding,'' \emph{science}, vol. 290, no. 5500, pp. 2323--2326, 2000.

\bibitem{belkin2002laplacian}
M.~Belkin and P.~Niyogi, ``Laplacian eigenmaps and spectral techniques for
  embedding and clustering,'' in \emph{Advances in neural information
  processing systems}, 2002, pp. 585--591.

\bibitem{jacob2014learning}
Y.~Jacob, L.~Denoyer, and P.~Gallinari, ``Learning latent representations of
  nodes for classifying in heterogeneous social networks,'' in
  \emph{Proceedings of the 7th ACM international conference on Web search and
  data mining}.\hskip 1em plus 0.5em minus 0.4em\relax ACM, 2014, pp. 373--382.

\bibitem{hofmann2001unsupervised}
T.~Hofmann, ``Unsupervised learning by probabilistic latent semantic
  analysis,'' \emph{Machine learning}, vol.~42, no. 1-2, pp. 177--196, 2001.

\bibitem{blei2003latent}
D.~M. Blei, A.~Y. Ng, and M.~I. Jordan, ``Latent dirichlet allocation,''
  \emph{Journal of machine Learning research}, vol.~3, no. Jan, pp. 993--1022,
  2003.

\bibitem{wang2017community}
X.~Wang, P.~Cui, J.~Wang, J.~Pei, W.~Zhu, and S.~Yang, ``Community preserving
  network embedding.'' 2017.

\bibitem{tang2015line}
J.~Tang, M.~Qu, M.~Wang, M.~Zhang, J.~Yan, and Q.~Mei, ``Line: Large-scale
  information network embedding,'' in \emph{Proceedings of the 24th
  International Conference on World Wide Web}.\hskip 1em plus 0.5em minus
  0.4em\relax International World Wide Web Conferences Steering Committee,
  2015, pp. 1067--1077.

\bibitem{li2016robust}
J.~Li, X.~Hu, L.~Wu, and H.~Liu, ``Robust unsupervised feature selection on
  networked data,'' in \emph{Proceedings of the 2016 SIAM International
  Conference on Data Mining}.\hskip 1em plus 0.5em minus 0.4em\relax SIAM,
  2016, pp. 387--395.

\bibitem{yang2015network}
C.~Yang, Z.~Liu, D.~Zhao, M.~Sun, and E.~Y. Chang, ``Network representation
  learning with rich text information.'' in \emph{IJCAI}, 2015, pp. 2111--2117.

\bibitem{huang2017accelerated}
X.~Huang, J.~Li, and X.~Hu, ``Accelerated attributed network embedding,'' in
  \emph{Proceedings of the 2017 SIAM International Conference on Data
  Mining}.\hskip 1em plus 0.5em minus 0.4em\relax SIAM, 2017, pp. 633--641.

\bibitem{zhang2017user}
D.~Zhang, J.~Yin, X.~Zhu, and C.~Zhang, ``User profile preserving social
  network embedding,'' in \emph{Proceedings of the 26th International Joint
  Conference on Artificial Intelligence}.\hskip 1em plus 0.5em minus
  0.4em\relax AAAI Press, 2017, pp. 3378--3384.

\bibitem{pan2016tri}
S.~Pan, J.~Wu, X.~Zhu, C.~Zhang, and Y.~Wang, ``Tri-party deep network
  representation,'' \emph{Network}, vol.~11, no.~9, p.~12, 2016.

\bibitem{Liao2017Attributed}
L.~Liao, X.~He, H.~Zhang, and T.~S. Chua, ``Attributed social network
  embedding,'' \emph{IEEE Transactions on Knowledge \& Data Engineering},
  vol.~PP, no.~99, pp. 1--1, 2017.

\bibitem{hardoon2004canonical}
D.~R. Hardoon, S.~Szedmak, and J.~Shawe-Taylor, ``Canonical correlation
  analysis: An overview with application to learning methods,'' \emph{Neural
  computation}, vol.~16, no.~12, pp. 2639--2664, 2004.

\bibitem{rosipal2005overview}
R.~Rosipal and N.~Kr{\"a}mer, ``Overview and recent advances in partial least
  squares,'' in \emph{International Statistical and Optimization Perspectives
  Workshop" Subspace, Latent Structure and Feature Selection"}.\hskip 1em plus
  0.5em minus 0.4em\relax Springer, 2005, pp. 34--51.

\bibitem{tenenbaum2000separating}
J.~B. Tenenbaum and W.~T. Freeman, ``Separating style and content with bilinear
  models,'' \emph{Neural computation}, vol.~12, no.~6, pp. 1247--1283, 2000.

\bibitem{srivastava2012learning}
N.~Srivastava and R.~Salakhutdinov, ``Learning representations for multimodal
  data with deep belief nets,'' in \emph{International conference on machine
  learning workshop}, vol.~79, 2012.

\bibitem{srivastava2014multimodal}
------, ``Multimodal learning with deep boltzmann machines,'' \emph{Journal of
  Machine Learning Research}, vol.~15, pp. 2949--2980, 2014.

\bibitem{kang2015learning}
C.~Kang, S.~Xiang, S.~Liao, C.~Xu, and C.~Pan, ``Learning consistent feature
  representation for cross-modal multimedia retrieval,'' \emph{IEEE
  Transactions on Multimedia}, vol.~17, no.~3, pp. 370--381, 2015.

\bibitem{xu2017learning}
X.~Xu, F.~Shen, Y.~Yang, H.~T. Shen, and X.~Li, ``Learning discriminative
  binary codes for large-scale cross-modal retrieval,'' \emph{IEEE Transactions
  on Image Processing}, vol.~26, no.~5, pp. 2494--2507, 2017.

\bibitem{ying2018graph}
R.~Ying, R.~He, K.~Chen, P.~Eksombatchai, W.~L. Hamilton, and J.~Leskovec,
  ``Graph convolutional neural networks for web-scale recommender systems,''
  \emph{arXiv preprint arXiv:1806.01973}, 2018.

\bibitem{defferrard2016convolutional}
M.~Defferrard, X.~Bresson, and P.~Vandergheynst, ``Convolutional neural
  networks on graphs with fast localized spectral filtering,'' in
  \emph{Advances in Neural Information Processing Systems}, 2016, pp.
  3844--3852.

\bibitem{kipf2016semi}
T.~N. Kipf and M.~Welling, ``Semi-supervised classification with graph
  convolutional networks,'' \emph{arXiv preprint arXiv:1609.02907}, 2016.

\bibitem{wu2018mining}
P.~Wu and L.~Pan, ``Mining application-aware community organization with
  expanded feature subspaces from concerned attributes in social networks,''
  \emph{Knowledge-Based Systems}, vol. 139, pp. 1--12, 2018.

\bibitem{cai2018three}
X.~Cai, J.~Han, W.~Li, R.~Zhang, S.~Pan, and L.~Yang, ``A three-layered
  mutually reinforced model for personalized citation recommendation,''
  \emph{IEEE Transactions on Neural Networks and Learning Systems}, 2018.

\bibitem{zhao2018learning}
W.~Zhao, S.~Tan, Z.~Guan, B.~Zhang, M.~Gong, Z.~Cao, and Q.~Wang, ``Learning to
  map social network users by unified manifold alignment on hypergraph,''
  \emph{IEEE Transactions on Neural Networks and Learning Systems}, 2018.

\bibitem{he2016deep}
K.~He, X.~Zhang, S.~Ren, and J.~Sun, ``Deep residual learning for image
  recognition,'' in \emph{Proceedings of the IEEE conference on computer vision
  and pattern recognition}, 2016, pp. 770--778.

\bibitem{charte2018practical}
D.~Charte, F.~Charte, S.~Garc{\'\i}a, M.~J. del Jesus, and F.~Herrera, ``A
  practical tutorial on autoencoders for nonlinear feature fusion: Taxonomy,
  models, software and guidelines,'' \emph{Information Fusion}, vol.~44, pp.
  78--96, 2018.

\bibitem{vincent2010stacked}
P.~Vincent, H.~Larochelle, I.~Lajoie, Y.~Bengio, and P.-A. Manzagol, ``Stacked
  denoising autoencoders: Learning useful representations in a deep network
  with a local denoising criterion,'' \emph{Journal of machine learning
  research}, vol.~11, no. Dec, pp. 3371--3408, 2010.

\bibitem{belkin2003laplacian}
M.~Belkin and P.~Niyogi, ``Laplacian eigenmaps for dimensionality reduction and
  data representation,'' \emph{Neural computation}, vol.~15, no.~6, pp.
  1373--1396, 2003.

\bibitem{hinton2006fast}
G.~E. Hinton, S.~Osindero, and Y.-W. Teh, ``A fast learning algorithm for deep
  belief nets,'' \emph{Neural computation}, vol.~18, no.~7, pp. 1527--1554,
  2006.

\bibitem{erhan2010does}
D.~Erhan, Y.~Bengio, A.~Courville, P.-A. Manzagol, P.~Vincent, and S.~Bengio,
  ``Why does unsupervised pre-training help deep learning?'' \emph{Journal of
  Machine Learning Research}, vol.~11, no. Feb, pp. 625--660, 2010.

\bibitem{simonyan2014very}
K.~Simonyan and A.~Zisserman, ``Very deep convolutional networks for
  large-scale image recognition,'' \emph{arXiv preprint arXiv:1409.1556}, 2014.

\bibitem{szegedy2015going}
C.~Szegedy, W.~Liu, Y.~Jia, P.~Sermanet, S.~Reed, D.~Anguelov, D.~Erhan,
  V.~Vanhoucke, A.~Rabinovich \emph{et~al.}, ``Going deeper with
  convolutions.''\hskip 1em plus 0.5em minus 0.4em\relax Cvpr, 2015.

\bibitem{zeiler2014visualizing}
M.~D. Zeiler and R.~Fergus, ``Visualizing and understanding convolutional
  networks,'' in \emph{European conference on computer vision}.\hskip 1em plus
  0.5em minus 0.4em\relax Springer, 2014, pp. 818--833.

\bibitem{fan2008liblinear}
R.-E. Fan, K.-W. Chang, C.-J. Hsieh, X.-R. Wang, and C.-J. Lin, ``Liblinear: A
  library for large linear classification,'' \emph{Journal of machine learning
  research}, vol.~9, no. Aug, pp. 1871--1874, 2008.

\end{thebibliography}
%

%

\begin{IEEEbiography}[{\includegraphics[width=1in,height=1.25in,clip,keepaspectratio]{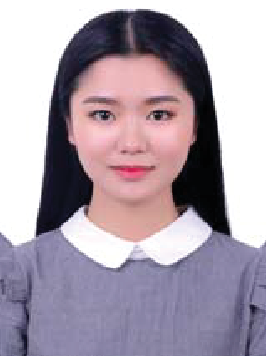}}]{Conghui Zheng}
Conghui Zheng is currently a PhD Student in Department of Electronic Engineering in Shanghai Jiao Tong University. She has received the B.S. degree in School of Information and Communication Engineering in Dalian  University of Technology, China, in 2016. Her research interests include data mining, neural network and representation learning.
\end{IEEEbiography}

\begin{IEEEbiography}[{\includegraphics[width=1in,height=1.25in,clip,keepaspectratio]{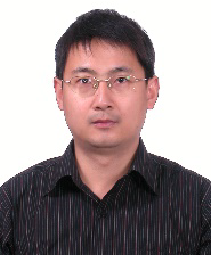}}]{Li Pan}
is a professor in the Cyber Security School of Shanghai Jiao Tong University. He received his Ph.D. degree in communication and information system from Shanghai Jiao Tong University, China, in 2002. Since 2002, he has been working in Shanghai Jiao Tong University. In 2008, he worked as a visiting scholar in IBM Research T.J. Watson Center, USA. In 2014, he worked as a visiting scholar in the computer department of Boston University, USA. His research interests focus on network management, system security and social computing. He published over 80 papers, 5 co-authored books on operating system and network security, and hold 10 granted patents. In 2007 and 2008, he was awarded by Shanghai Rising-Star Program for Young Scientists in China, and Foundation for the Shanghai Talents in China respectively. In 2013, he was elected for the Shanghai Subject Chief Scientist in China.
\end{IEEEbiography}

\begin{IEEEbiography}[{\includegraphics[width=1in,height=1.25in,clip,keepaspectratio]{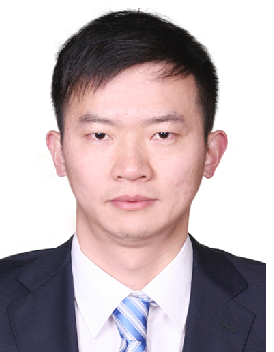}}]{Peng Wu}
received the B.S. degree in School of Information Science and Engineering in Southeast University, Nanjing, China, in 2012, and received the Ph.D. degree in Department of Electronic Engineering in Shanghai Jiao Tong University, Shanghai, China, in 2018. Since 2018, he has been working in Shanghai Jiao Tong University. His research interests include social computing, data mining and computational intelligence.
\end{IEEEbiography}

\section*{Appendices}
In order to better compare the close curves in Figure 6-7, numeric results are provided in Table 9-11. Numbers in bold represent the best result in each column.

\begin{table*}[!h]
\normalsize
\renewcommand{\arraystretch}{1.3}
\caption{AUC of Attribute Prediction}
\label{T9}
\centering
\begin{tabular}{|c|c|c|c|c|c|c|}
\hline
     &Test Ratio & 0.05    & 0.15    & 0.25    & 0.35     & 0.45    \\
\hline
               & LE & 0.713 & 0.700 & 0.691 & 0.681 & 0.667   \\
   & node2vec & 0.626 & 0.620 & 0.613 & 0.606 & 0.597  \\
cora & SDNE & 0.701 & 0.686 & 0.677 & 0.665 & 0.652  \\
       & AANE &  0.630 & 0.620 & 0.613 & 0.605 & 0.597  \\
       & ASNE  & 0.710  & 0.688 & 0.676 & 0.660 & 0.643  \\
      & MDNE & \textbf{0.744}& \textbf{0.726} & \textbf{0.714} & \textbf{0.698} & \textbf{0.680} \\
\hline
               & LE & 0.700 & 0.691 & 0.681 & 0.671 & 0.657  \\
   & node2vec & 0.629 & 0.623 & 0.615 & 0.606 & 0.595  \\
citeseer & SDNE & 0.715 & 0.704 & 0.693 & 0.680 & 0.664  \\
       & AANE &  0.634 & 0.626 & 0.618 & 0.608 & 0.597  \\
       & ASNE  & 0.729  & 0.710 & 0.691 & 0.671 & 0.649  \\
      & MDNE & \textbf{0.776}& \textbf{0.759} & \textbf{0.741} & \textbf{0.721} & \textbf{0.697} \\
\hline
               & LE & 0.803 & 0.796 & 0.793 & 0.784 & 0.778  \\
UNC & SDNE & 0.865& 0.858 & 0.852 & 0.844 & 0.837  \\
       & AANE & 0.780  & 0.771 & 0.768 & 0.754 & 0.747  \\
       & ASNE  & 0.780  & 0.773 & 0.770 & 0.760 & 0.751  \\
      & MDNE & \textbf{0.881}& \textbf{0.874} & \textbf{0.868} & \textbf{0.860} & \textbf{0.853} \\
\hline
               & LE & 0.800 & 0.795 & 0.790 & 0.786 & 0.780  \\
Oklahoma & SDNE &0.826&0.821&0.816&0.810&0.806  \\
       & AANE & 0.759 & 0.755 & 0.751 & 0.745 & 0.741  \\
       & ASNE  & 0.870 & 0.863 & 0.857 & 0.851 & 0.844  \\
      & MDNE & \textbf{0.872}& \textbf{0.863} & \textbf{0.857} & \textbf{0.851} & \textbf{0.845} \\
\hline
\end{tabular}
\end{table*}

\begin{table*}[!h]
\normalsize
\renewcommand{\arraystretch}{1.3}
\caption{Macro-F1 on cora and citeseer for Classification}
\label{T10}
\centering
\begin{tabular}{|c|c|c|c|c|c|c|c|c|c|c|}
\hline
     &Test Ratio & 0.1    & 0.2     & 0.3     & 0.4      & 0.5     & 0.6     & 0.7      & 0.8      & 0.9    \\
\hline
               & LE & 0.716 & 0.700 & 0.701 & 0.715 & 0.688 & 0.681 & 0.655 & 0.621 & 0.535 \\
   & node2vec & 0.139 & 0.139 & 0.133 & 0.133 & 0.137 & 0.139 & 0.137 & 0.140 & 0.142  \\
cora & SDNE & 0.525 & 0.504 & 0.505 & 0.503 & 0.474 & 0.456 & 0.417 & 0.371 & 0.341  \\
       & AANE &  0.127 & 0.136 & 0.133 & 0.134 & 0.133 & 0.140 & 0.140 & 0.146 & 0.149  \\
       & ASNE  & 0.444  & 0.429 & 0.406 & 0.421 & 0.437 & 0.410 & 0.419 & 0.350 & 0.343  \\
      & MDNE & \textbf{0.790}& \textbf{0.802} & \textbf{0.779} & \textbf{0.748} & \textbf{0.738} & \textbf{0.726} & \textbf{0.719} & \textbf{0.688} & \textbf{0.660} \\
\hline
               & LE & 0.519 & 0.497 & 0.504 & 0.496 & 0.495 & 0.481 & 0.467 & 0.452 & 0.408 \\
   & node2vec & 0.166 & 0.176 & 0.168 & 0.172 & 0.170 & 0.171 & 0.173 & 0.176 & 0.171  \\
citeseer & SDNE & 0.377 & 0.375 & 0.362 & 0.369 & 0.366 & 0.353 & 0.345 & 0.324 & 0.273  \\
       & AANE &  0.161 & 0.173 & 0.177 & 0.172 & 0.176 & 0.176 & 0.173 & 0.176 & 0.169  \\
       & ASNE  & 0.328  & 0.325 & 0.323 & 0.311 & 0.324 & 0.306 & 0.317 & 0.315 & 0.286  \\
      & MDNE & \textbf{0.634}& \textbf{0.648} & \textbf{0.629} & \textbf{0.624} & \textbf{0.623} & \textbf{0.609} & \textbf{0.583} & \textbf{0.509} & \textbf{0.473} \\
\hline
\end{tabular}
\end{table*}

\begin{table*}[!h]
\normalsize
\renewcommand{\arraystretch}{1.3}
\caption{Micro-F1 on cora and citeseer for Classification}
\label{T11}
\centering
\begin{tabular}{|c|c|c|c|c|c|c|c|c|c|c|}
\hline
     &Test Ratio & 0.1    & 0.2     & 0.3     & 0.4      & 0.5     & 0.6     & 0.7      & 0.8      & 0.9    \\
\hline
               & LE & 0.725 & 0.710 & 0.708 & 0.721 & 0.696 & 0.691 & 0.666 & 0.636 & 0.555 \\
   & node2vec & 0.163 & 0.163 & 0.157 & 0.158 & 0.161 & 0.163 & 0.161 & 0.163 & 0.165  \\
cora & SDNE & 0.541 & 0.526 & 0.528 & 0.524 & 0.497 & 0.485 & 0.446 & 0.395 & 0.376  \\
       & AANE &  0.255 & 0.257 & 0.251 & 0.241 & 0.233 & 0.226 & 0.222 & 0.216 & 0.208  \\
       & ASNE  & 0.488  & 0.479 & 0.447 & 0.467 & 0.472 & 0.458 & 0.459 & 0.378 & 0.389  \\
      & MDNE & \textbf{0.807}& \textbf{0.815} & \textbf{0.798} & \textbf{0.774} & \textbf{0.761} & \textbf{0.748} & \textbf{0.743} & \textbf{0.711} & \textbf{0.687} \\
\hline
               & LE & 0.534 & 0.499 & 0.517 & 0.515 & 0.501 & 0.485 & 0.471 & 0.468 & 0.416 \\
   & node2vec & 0.179 & 0.184 & 0.182 & 0.184 & 0.181 & 0.181 & 0.183 & 0.185 & 0.183  \\
citeseer & SDNE & 0.439 & 0.427 & 0.412 & 0.419 & 0.409 & 0.397 & 0.382 & 0.353 & 0.295  \\
       & AANE &  0.196 & 0.210 & 0.214 & 0.206 & 0.204 & 0.203 & 0.198 & 0.203 & 0.192  \\
       & ASNE  & 0.355  & 0.360 & 0.359 & 0.341 & 0.355 & 0.335 & 0.347 & 0.341 & 0.308  \\
      & MDNE & \textbf{0.691}& \textbf{0.700} & \textbf{0.691} & \textbf{0.678} & \textbf{0.673} & \textbf{0.656} & \textbf{0.623} & \textbf{0.545} & \textbf{0.502} \\
\hline
\end{tabular}
\end{table*}






\end{document}